\documentclass[10pt,twocolumn]{article}

\usepackage[a4paper,margin=0.75in,columnsep=0.25in]{geometry}
\usepackage[utf8]{inputenc}
\usepackage[T1]{fontenc}
\usepackage{newtxtext,newtxmath}
\usepackage{microtype}
\usepackage{booktabs}
\usepackage{tabularx}
\usepackage{array}
\usepackage{longtable}
\usepackage{graphicx}
\usepackage{xcolor}
\usepackage{enumitem}
\usepackage{tikz}
\usepackage{pifont}
\usepackage{pgfplots}
\pgfplotsset{compat=1.17}
\usetikzlibrary{positioning,arrows.meta}
\usepackage[numbers,sort&compress]{natbib}
\usepackage[hidelinks]{hyperref}
\usepackage{cleveref}
\usepackage{caption}
\graphicspath{{figures/}}

\newcommand{\pmax}{p_{\max}}
\newcommand{\Hn}{\tilde{H}}
\newcommand{\TV}{\mathrm{TV}}
\newcommand{\sece}{\mathrm{smECE}}

\title{\Large\bfseries
Beyond Answer Confidence: \\ A Controlled Audit of Self-Knowledge in a Black-Box Decision  Model
}
\author{Sharath M Shankaranarayana, Davor Runje and Jan Jannink\\
\small Synthpop.AI\\
\small\texttt{sharath@synthpop.ai}, \texttt{davor@synthpop.ai},
\texttt{jan@synthpop.ai}}
\date{}

\begin{document}
\maketitle

\begin{abstract}
Decision models return probabilities intended for routing, abstention and
automated action. Calibration makes those probabilities useful on average,
but does not establish whether low confidence reflects chance or missing
knowledge, nor whether confidence falls when a model moves beyond what it
knows. We audit this distinction in Jev, a decision model, with over 15 public datasets and 6 generated task families,
with paired interventions that vary the information supplied for a fixed item.
Jev's confidence is calibrated on familiar closed-choice tasks but fails as an
indicator of missing knowledge: with no answer-relevant information it assigns
up to 0.80 to a salient option, and on news beyond an observed knowledge
boundary it exceeds accuracy by 0.21--0.33, a gap that recalibration on
earlier months does not close. Targeted yes/no questions give sharper
readouts of the case: whether an outcome is settled (AUROC 1.00) and whether
the evidence suffices (0.95, against 0.85 for confidence on the same items).
Asking whether Jev knows the answer appears to flag fabricated entities and
post-boundary news (0.91), but with realistic names or with dates removed it
shows no advantage over answer uncertainty. Black-box knowledge audits
therefore need explicit controls for surface cues. Code:
\url{https://github.com/Syntheme/beyond-answer-confidence}.
\end{abstract}

\section{Introduction}
\label{sec:intro}

Recently, there has been a surge in interest in decision-focused AI models.
This class of models returns typed probability distributions rather than
free-form text. These probabilities are intended to support consequential
decisions: accept an answer, abstain, route the request to another system, or
trigger human review. Their usefulness therefore depends not only on average
calibration, but on whether they remain informative when the model lacks the
knowledge needed for a particular case.

\begin{figure}[t]
\centering
\begin{tikzpicture}[
  font=\scriptsize,
  box/.style={draw,rounded corners=2pt,align=center,inner sep=2.5pt,minimum height=1.15cm},
  outbox/.style={draw,rounded corners=2pt,align=left,inner sep=2.5pt,text width=3.55cm,minimum height=2.5cm},
  arr/.style={-{Stealth[length=1.5mm]},semithick}]
\node[box,fill=blue!6,text width=2.35cm] (k) {\textbf{Change what Jev is told}\\ examples, facts,\\ paragraphs; item fixed};
\node[box,fill=orange!8,text width=2.35cm,right=0.12cm of k] (c) {\textbf{State the chance}\\ stated odds, dice,\\ coins};
\node[box,fill=gray!6,dashed,text width=2.35cm,right=0.12cm of c] (o) {\textbf{Probe the boundary}\\ made-up entities,\\ recent news;\\ cue controls};
\node[draw,rounded corners=2pt,fill=gray!15,align=center,text width=7.6cm,below=0.35cm of c,inner sep=3pt] (m)
  {\textbf{Jev}: one call with a JSON state and typed questions (Choice, yes/no, Score)};
\draw[arr] (k.south) -- (k.south |- m.north);
\draw[arr] (c.south) -- (m.north);
\draw[arr] (o.south) -- (o.south |- m.north);
\node[outbox,fill=red!4,below=0.35cm of m.south west,anchor=north west] (a)
  {\textbf{Confidence} (top probability)\\[2pt]
   \textcolor{green!45!black}{\ding{51}} calibrated on familiar\\ \phantom{\ding{51}} closed-choice tasks\\
   \textcolor{red!70!black}{\ding{55}} no information: backs a\\ \phantom{\ding{55}} salient guess (up to 0.80)\\
   \textcolor{red!70!black}{\ding{55}} past the knowledge boundary:\\ \phantom{\ding{55}} \emph{more} confident, sensitive\\ \phantom{\ding{55}} to dates; confidence-only\\ \phantom{\ding{55}} recalibration leaves a gap};
\node[outbox,fill=green!6,below=0.35cm of m.south east,anchor=north east] (f)
  {\textbf{Targeted follow-up questions}\\[2pt]
   \textcolor{green!45!black}{\ding{51}} \texttt{settled}? outcome not yet\\ \phantom{\ding{51}} fixed (AUROC 1.00)\\
   \textcolor{green!45!black}{\ding{51}} \texttt{enough}? evidence missing\\ \phantom{\ding{51}} from the context (0.95)\\
   \textcolor{red!70!black}{\ding{55}} \texttt{known}? reads name form\\ \phantom{\ding{55}} and dates; with those cues\\ \phantom{\ding{55}} controlled, no advantage\\ \phantom{\ding{55}} over answer uncertainty};
\draw[arr] (m.south -| a.north) -- (a.north);
\draw[arr] (m.south -| f.north) -- (f.north);
\end{tikzpicture}
\caption{Overview. We change what Jev is told for a fixed item, state the
chance of an outcome, or probe the knowledge boundary (dashed:
observational; cue controls use realistic names and remove dates), and read
two outputs from the same call. Confidence predicts correctness on familiar
closed-choice tasks but does not show what Jev does not know; questions about
the case detect gaps in it, while ``do you know?'' mostly reads surface cues
(\cref{tab:gaps}, \cref{fig:cues}).}
\label{fig:overview}
\end{figure}
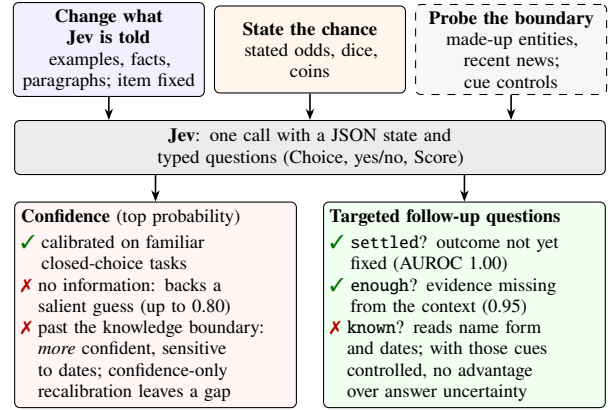

Calibration alone does not provide that guarantee. A calibrated probability
of 0.5 may describe a fair coin or an unknown fact, even though the first
reflects chance and the second missing knowledge. More generally, the split
of a first-order predictive distribution into aleatoric and epistemic
uncertainty is not determined by that distribution
\citep{bengs2022pitfalls,wimmer2023quantifying,ahdritz2025provable}.
Consequently, epistemic uncertainty cannot be identified by inspecting one
output distribution in isolation.

It can, however, be audited behaviourally. We can hold an item and its
ambiguity fixed while varying the information available to the model; compare
fixed unknown facts with future random outcomes having the same ideal answer
distribution; and ask separate questions about whether an outcome is settled,
whether the supplied evidence suffices, or whether the model is familiar with
the subject. These interventions test whether uncertainty responds to missing
information without assuming that the model internally represents an
epistemic--aleatoric decomposition.

We apply this audit to Jev, a decision model launched recently. Jev returns probabilities over allowed answers and a derived
\texttt{confidence} score. Its authors state that higher confidence predicts
higher accuracy and attributes this behaviour to reinforcement learning for
calibrated decisions
\citep{typesafe2026launch,typesafe2026confidence}. We distinguish that
calibration claim from a stronger operational question: \emph{does Jev's
top-answer probability remain informative when the knowledge required to
answer is absent?}

\paragraph{Main finding.}
Jev's confidence predicts correctness on familiar material but does not
identify why a case is uncertain. It remains high beyond an observed
knowledge boundary and selects salient answers when the input provides no
basis for choosing one. Targeted questions asked on the same call give direct
readouts that confidence provides only indirectly or with the wrong
orientation: whether the outcome is settled, whether the evidence is
sufficient, and whether the subject is familiar
(\cref{fig:overview}, \cref{tab:gaps}). Controls for surface cues separate
these: the first two, which describe the case, hold on the tested contrasts;
the third, which concerns the model's own knowledge, is largely explained by
unfamiliar-looking names and recent dates: it provides no evidence of
memory-gap detection beyond confidence once the measured cues are controlled. A model's ability to assess the case
it is shown should therefore not be mistaken for an ability to identify gaps
in its own knowledge.

\begin{table*}[t]
\centering
\small
\caption{Each contrast, what it was intended to measure, and what the controls
show it measures. AUROC for the follow-up question and, on the same items,
for answer uncertainty $1-\pmax$, or for confidence $\pmax$ where that is the
orientation that separates the contrast (as marked); ``wordings'' is the range over
paraphrased, reversed and adversarial versions. The contrasts are different tasks, so their AUROCs are not directly
comparable. Both news rows use the same 3{,}360 questions (1{,}680 per period);
wordings for \texttt{known} were measured on all real subjects (AUROC 0.91
there). Fabricated-versus-real status and period are proxies for what the
model knows, not direct labels of it. The cue-controlled rows (look-alike
names, dates removed) are exploratory analyses added after the main studies.}
\label{tab:gaps}
\setlength{\tabcolsep}{4pt}
\begin{tabular}{@{}>{\raggedright\arraybackslash}p{0.21\textwidth} >{\raggedright\arraybackslash}p{0.13\textwidth} >{\raggedright\arraybackslash}p{0.22\textwidth} >{\raggedright\arraybackslash}p{0.14\textwidth} >{\raggedright\arraybackslash}p{0.07\textwidth} r r@{}}
\toprule
Ground-truth contrast & Intended to measure & Validated interpretation & Answer uncertainty & Question & AUROC & Wordings \\
\midrule
Fixed but unknown fact vs.\ future chance event & outcome settled & outcome settled, also inferred from tense alone (1.00) & 0.96, only because Jev concentrates on the mode for chance events; ideal distributions identical & \texttt{settled} & 1.00 & 0.99--1.00 \\
Complete vs.\ incomplete evidence, same paragraph count & evidence sufficient & evidence sufficient, beyond length features (0.60 $\to$ 0.94) & 0.85 (confidence, higher = complete) & \texttt{enough} & 0.95 & 0.95--0.97 \\
\midrule
Pseudo-word made-up vs.\ obscure real person (same relations) & model's knowledge & name form: text alone 1.00 & 0.78 & \texttt{known} & 0.88 & 0.87--0.97 \\
Look-alike made-up vs.\ obscure real person & model's knowledge & equivalent to answer uncertainty ($\Delta=-0.02$) & 0.76 & \texttt{known} & 0.74 & --- \\
News past vs.\ before the boundary & model's knowledge & explicit dates: date alone 0.997 & 0.385 (0.615 reversed) & \texttt{known} & 0.91 & --- \\
\quad same, answer option & & & & ``not known'' & 0.89 & --- \\
News past vs.\ before, dates removed & model's knowledge & no demonstrated advantage over answer uncertainty ($\Delta=+0.02$) & 0.56 & \texttt{known} & 0.58 & --- \\
\bottomrule
\end{tabular}
\end{table*}

\paragraph{Contributions.}
\begin{itemize}[leftmargin=*,itemsep=1pt,topsep=2pt]
\item \textbf{Controlled information interventions reveal where confidence
  fails.} Jev's top-answer probability is calibrated on familiar closed-choice
  tasks, but favors salient options without answer-relevant information and
  becomes overconfident beyond an observed knowledge boundary; recalibrating
  on earlier months leaves a large post-boundary gap.

\item \textbf{Surface-cue controls expose the limits of apparent
  self-knowledge.} Asking whether Jev knows the answer looks like a strong
  readout on the original contrasts, but text-only baselines, realistic names
  and date removal show that name form and dates carry most of that signal.

\item \textbf{Questions about the case remain useful.} Asking whether an
  outcome is settled and whether the supplied evidence suffices gives direct,
  correctly oriented readouts of the tested contrasts (confidence separates
  them only through mode-seeking or less sharply); an explicit ``not known''
  option helps when many questions lie beyond the boundary.

\item \textbf{A behavioural audit for black-box decision models.}
  We combine paired interventions that vary supplied information while
  holding the item fixed, generated tasks with known ideal distributions,
  and separate queries about the case, evidence and model familiarity. The
  audit comprises about 575{,}000 calls across 15 public datasets and 6
  generated task families, with text-only baselines and cue-controlled
  variants for the follow-up questions.
\end{itemize}

\section{Related work}
\label{sec:related}

\paragraph{Epistemic and aleatoric uncertainty.}
Epistemic uncertainty is the learner's lack of knowledge of $p^*(y\mid x)$,
aleatoric uncertainty its intrinsic randomness
\citep{hullermeier2021aleatoric}. The entropy/mutual-information split is
contested \citep{wimmer2023quantifying,schweighofer2025information,bickfordsmith2025rethinking},
second-order losses do not incentivise faithful epistemic estimates
\citep{bengs2022pitfalls,jurgens2024epistemic}, provable decomposition needs
higher-order supervision \citep{ahdritz2025provable,johnson2024experts}, and
even white-box methods disentangle the two poorly
\citep{mucsanyi2024benchmarking}. A model that returns one distribution per
query offers nothing to decompose, so we test behaviour.

\paragraph{Black-box tests.}
We build on sensitivity to irrelevant context \citep{shi2023distracted};
knowledge boundaries via popularity \citep{mallen2023popqa}, fabricated
entities and unanswerable questions
\citep{yin2023selfaware,amayuelas2024knowledge,kirichenko2025abstentionbench};
temporal cutoffs \citep{dai2025dailyoracle}; self-evaluation
\citep{kadavath2022language,tian2023just,xiong2024can}; iterative prompting
\citep{abbasiyadkori2024believe}; and human label distributions
\citep{nie2020chaosnli,baan2022stop}.

\paragraph{Closest work.}
LM probabilities deviate from stated odds in coin and dice contexts
\citep{lovering2025numeric}; token and verbalised probabilities can disagree
\citep{expressdoubts2025}; and models that describe a fair coin do not behave
like one \citep{flipping2025}. Concurrent work tests fabricated entities and
post-cutoff events across many LLMs \citep{senol2026dkwtdk2}. For Jev,
early third-party tests compared its multiple-choice and yes/no outputs on
hidden fair draws \citep{hayashi2026dice}, reported its fair-die preference
\citep{typellm2026die}, and measured calibration on intent benchmarks
\citep{assay001}. Concurrent papers evaluate Jev as an evaluation judge,
where its confidence tracks correctness well enough to route uncertain cases
to a stronger model \citep{li2026jevjudge}, and for semantic choices in
scientific pipelines \citep{deng2026jevscience}. Those works ask whether
confidence predicts correctness; we ask whether it reports missing knowledge. We differ in holding the aleatoric part fixed by design, in
tasks with a known ideal for both kinds of uncertainty, and in separating what
different follow-up questions measure.

\section{Setup}
\label{sec:setup}

\paragraph{The model.}
Jev takes a JSON \emph{state}, natural-language instructions and an allowed
answer type \citep{typesafe2026api}: a \textbf{Choice} returns probabilities
over up to 255 named options, a \textbf{Noul} (Jev's yes/no type) returns
$P(\text{yes})$, and a \textbf{Score} returns probabilities over at most ten
ordered levels (\cref{app:example} shows a request and response). The API
exposes only aliases; we requested \texttt{jev-latest}, and every response
reported \texttt{jev-1.13.0} (collection 26--30 September 2026; the surface-cue controls and verification variants were run on 30 September). At the time,
input cost \$0.042 per million tokens and output was free; our cost figures
are estimates from token counts, not invoices. The vendor's documentation
acknowledges that answers to related questions need not be coherent
\citep{typesafe2026jaggedness}. Jev is not deterministic (replicate SD $\approx$
0.04 near $p=0.5$), probabilities are quantised to 0.01, and several questions
in one call do not change answers beyond replicate noise. We queried every
cell three times and analyse the mean distribution
$\bar q(y)=\tfrac13\sum_{r=1}^{3}q_r(y)$, whose top option is Jev's answer and
whose top probability is its confidence; across 500 random
selections of one call per item, the 95\% SmoothECE ranges stay within 0.010
of the three-call estimates in all 36 calibration sets (\cref{app:single}).

\paragraph{Metrics.} By Jev's \emph{confidence} we mean the probability it
assigns to its chosen answer, the top probability $\pmax$. The API's
\texttt{confidence} field is a rescaling of the same quantity (close to
$(K\pmax-1)/(K-1)$ for $K$ options, 0 at uniform and 1 at certainty), so it
carries the same information; we use $\pmax$ because calibration needs the
probability scale. We report top-label SmoothECE
\citep{blasiok2024smooth}, the mean gap between confidence and accuracy,
AUROC for error or case detection, normalised entropy $\Hn=H/\log K$, and total
variation $\TV$ to an ideal distribution. Intervals are 95\% item-clustered
bootstrap intervals (month-clustered for dated news). We call confidence
``calibrated'' when its estimated top-label calibration error is low
(SmoothECE $\le 0.05$) on the closed-choice form of a task; this does not imply
classwise or subgroup calibration.

\paragraph{Five checks.}
Let $q(\cdot\mid x,\mathcal K)$ be the answer distribution for input $x$ with
side information $\mathcal K$. Uncertainty that reports missing knowledge
should show: \textbf{(C1)} \emph{reducibility}---evidence that narrows the
correct answers lowers entropy; \textbf{(C2)} \emph{honest
ignorance}---if $\mathcal K$ says nothing about the answer, $q$ is near uniform
over the options not ruled out; \textbf{(C3)} \emph{invariance to
non-evidence}---text independent of the answer, declared uninformative where
needed, leaves $q$ unchanged; \textbf{(C4)} \emph{calibration at every
knowledge level}; and \textbf{(C5)} a \emph{separate readout}---some other
query distinguishes missing knowledge from chance. The checks assume our
controlled, symmetric task constructions: evidence that provably narrows the
correct answers, options exchangeable under no information, and tasks with a
defined answer. They form an audit checklist, not a definition: passing them does not prove that a model knows
what it does not know; failures identify limitations in the tested
operational behaviour, and passing the checklist does not establish access to
a general epistemic state.

\paragraph{Studies.}
\Cref{fig:overview} summarises the design. The main studies are:
\begin{itemize}[leftmargin=*,itemsep=1pt,topsep=2pt]
\item \emph{Intent dose--response.} Banking77 \citep{casanueva2020efficient}
  (77 intents, 770 test items) and CLINC150 \citep{larson2019evaluation}
  (150 intents, 1{,}500 items), with intents shown as opaque codes
  (\texttt{INTENT\_00}, \ldots) that carry nothing, 1--8 example utterances,
  the intent name, name plus 8 examples, matched-length irrelevant text, or
  examples swapped across codes. Utterance and label stay fixed.
\item \emph{Knowledge$\times$chance scenarios.} 480 generated 4-option
  questions about fabricated entities in which the answer is stated, unknown
  with 0--2 options ruled out, or left to a future random draw with stated
  chances; the ideal distribution is known for every cell.
\item \emph{Knowledge boundary.} PopQA by subject popularity
  \citep{mallen2023popqa}, 1{,}600 fabricated-entity questions in PopQA's
  templates, and Daily Oracle yes/no news questions from January 2020 to July
  2026
  \citep{dai2025dailyoracle}.
\item \emph{Evidence in context.} HotpotQA comparison questions
  \citep{yang2018hotpotqa} (two options) with zero, one or two supporting
  paragraphs,
  Quizbowl \citep{rodriguez2019quizbowl} clue by clue, and 600 generated
  policy cases with a fact edited, deleted, contradicted or joined by an
  irrelevant one.
\item \emph{Random devices and answer formats}: dice, coins and cards over
  all option orders where feasible (sampled orders for the sum of two dice),
  and stated odds asked as one Choice or as one yes/no
  question per outcome.
\end{itemize}
Standard benchmarks, out-of-scope detection, an open decision model
(GLiNER2.5-Decide) and a deep ensemble complete the study
(\cref{app:more,app:datasets}). Free-text benchmarks were converted to
four-option Choice. Paired input changes support causal statements about how
Jev's outputs respond (though a change may also alter length or salience);
comparisons across popularity, dates and benchmarks are observational and
serve as convergent evidence.

\paragraph{Follow-up questions.} Each call also asks yes/no questions:
\texttt{settled} (``is the answer already fixed, so that someone with more
information could know it?''), \texttt{enough} (``do the paragraphs contain
enough information to answer for certain?'') and \texttt{known} (``do you know
the answer for certain, rather than having to guess?''; for dated news, ``do
you know for certain how this turned out?'').

\section{Results}
\label{sec:results}

\subsection{Confidence: trustworthy when familiar, blind at the edge}
\label{sec:answer}

\begin{table}[t]
\centering
\small
\setlength{\tabcolsep}{3.5pt}
\caption{Intent dose--response: accuracy, mean confidence and SmoothECE by
what each intent code carries. With swapped examples, accuracy and SmoothECE
are scored against the original labels; the near-zero accuracy shows that Jev
follows the swapped mapping, not a calibration failure.}
\label{tab:intent}
\begin{tabular}{@{}l rrr rrr@{}}
\toprule
& \multicolumn{3}{c}{Banking77 ($K$=77)} & \multicolumn{3}{c}{CLINC150 ($K$=150)} \\
\cmidrule(lr){2-4}\cmidrule(lr){5-7}
Code carries & acc & $\pmax$ & $\sece$ & acc & $\pmax$ & $\sece$ \\
\midrule
nothing & .010 & .320 & .310 & .008 & .363 & .355 \\
1 example & .681 & .828 & .170 & .911 & .904 & .024 \\
8 examples & .899 & .927 & .037 & .979 & .974 & .016 \\
name & .821 & .893 & .084 & .926 & .933 & .025 \\
name + 8 & .909 & .939 & .039 & .981 & .978 & .015 \\
irrelevant text & .019 & .312 & .293 & .007 & .226 & .219 \\
swapped examples & .001 & .932 & .904 & .001 & .974 & .957 \\
\bottomrule
\end{tabular}
\end{table}

\textbf{Familiar material.} Evidence reduces uncertainty
(\cref{tab:intent}): entropy falls sharply from no information to one example
per intent, and with swapped examples Jev follows the swapped mapping (90\% /
98\% of items), so it uses the evidence it is given. Confidence is calibrated
on familiar closed-choice tasks---CLINC150 at every informative level, with
examples or names (SmoothECE 0.015--0.025), PopQA in every popularity
quintile (0.020--0.033), MMLU-Redux (0.016), TriviaQA (0.031)---and useful for
abstention: answering only at $\pmax\ge0.9$ keeps 8{,}132 of 9{,}960 TriviaQA
questions with 20 errors.

\begin{table}[t]
\centering
\small
\setlength{\tabcolsep}{3pt}
\caption{Top probability when nothing in the input identifies the answer.
Where a hidden answer exists, the ideal is uniform over the options not ruled
out. $^\dagger$No valid answer exists (false premise); the options are real
people, so 0.25 is a uniform forced-choice reference, not a calibrated target.
$^\S$The outcome is not yet decided; the contenders are interchangeable
made-up names by construction, so 0.25 is the symmetric value.}
\label{tab:ignorance}
\begin{tabular}{@{}l r r@{}}
\toprule
Setting & Ideal & $\pmax$ \\
\midrule
Opaque codes (Banking77/CLINC150) & .013/.007 & .32/.36 \\
\quad same, open model & .013/.007 & .017/.010 \\
Fixed but unknown generated fact & .25 & .49 \\
Fabricated entities (PopQA templates) & .25$^\dagger$ & .52 \\
Winner of a made-up future event & .25$^\S$ & .76 \\
Fair die, all 720 option orders & .17 & .80 \\
\bottomrule
\end{tabular}
\end{table}

\textbf{No information: a salient guess} (\cref{tab:ignorance}). With opaque
codes only, accuracy is at chance but Jev puts a third of its mass on the first
numbered code; for a fair die it puts 0.80 on ``one'' in every option order.
The open GLiNER2.5-Decide model \citep{zaratiana2024gliner,fastino2025gliner25}
is almost exactly uniform on the same inputs, so this is not intrinsic to
decision models. Uninformative cues also move the answer: a guess ``made with
no information'' raises the guessed option by 0.65.

\begin{figure*}[t]
\centering
\includegraphics[width=\textwidth]{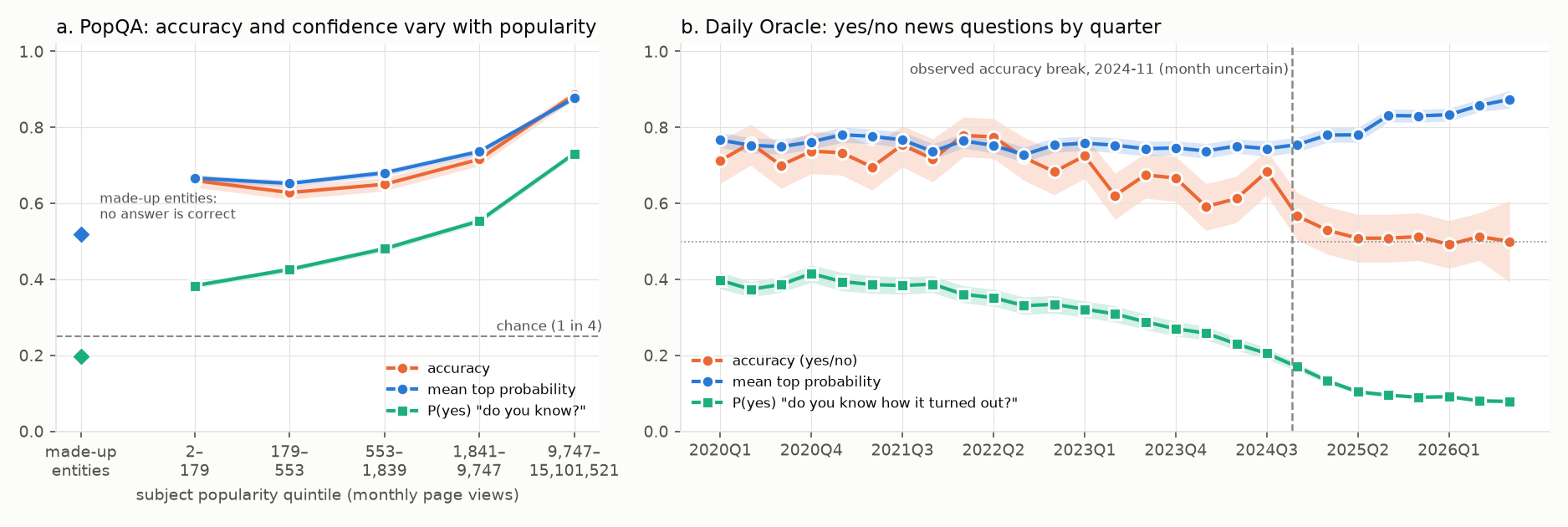}
\caption{Knowledge boundary. (a) PopQA: confidence tracks accuracy across
popularity quintiles, but fabricated entities still get 0.52 on a top option,
while $P(\text{known})$ drops to 0.20. (b) Daily Oracle yes/no questions by
quarter: after the observed accuracy break (change point, around late 2024) accuracy falls to
chance while confidence \emph{rises}; $P(\text{known})$ declines steadily, a
signal carried mostly by the dates in the questions (\cref{sec:followups}).}
\label{fig:boundary}
\end{figure*}

\textbf{The knowledge boundary: a confident default} (\cref{fig:boundary};
observational evidence). For fabricated entities, where no option is
correct, Jev still puts 0.52 on one option (a uniform forced-choice reference
would be 0.25). On yes/no
news past an observed accuracy change point around late 2024, Jev answers
``no'' to 92\% of questions, against 61\% before, although the true share of
``no'' is 51--52\% in both periods (51.2\% of 4{,}640 before, 51.5\% of
1{,}680 after), and its confidence \emph{rises} from 0.75
to 0.82. The dates in the questions (98.7\% contain a year) drive much of
this: with every temporal expression removed, post-boundary ``no'' answers fall
from 92\% to 67\%. Some removals change the proposition (``Will the Fed cut
rates by the end of March 2024?'' loses its deadline) or remove which event is
meant, so this is a change in responses, not a validated change in accuracy.
Among questions whose original answer is ``yes'' (1{,}648), removing dates cut
post-boundary ``no'' answers from 92\% to 57\%, against 41\% to 37\% before
the boundary; but in a 240-pair sample provisionally annotated by an AI
assistant under a written protocol, 30 of the 112 ``yes'' questions lost event
identity, so these are not all answer-safe. The pairs judged
answer-preserving show the same direction (92\% to 65\% after, $n=37$; 39\% to
35\% before, $n=46$; \cref{app:dates}). Removing dates also removes the
specificity that identifies an event, so these edits do not separate
superficial date-reading from temporal knowledge retrieval. The question's own date matters, not the
date of asking: stating today's date as just after the event leaves the
post-boundary ``no'' share at 91\%. The preference is not tied to the word
``no'': Jev verified both a proposed ``yes'' and a proposed ``no'' for every
item, with both a positive (``is the answer correct?'') and a negative (``is it
wrong?'') wording. The two wordings agree on 95--96\% of items, and past the
boundary both imply ``yes'' for only 8--10\% of questions (true share 48\%); a
preference for the word ``no'' would make the negative wording accept both
proposals, which happens for 0.1\%. This is behavioural evidence of a default
towards ``it did not happen'', not proof of a belief. Shifts between broad news categories alone do
not explain the drop, since accuracy falls within every category. The change point is an accuracy break, not a verified training
cutoff. On these yes/no questions Jev's log loss (0.71) is worse than always
answering 50/50 (0.69). Material built to be novel shows the
same pattern (MMLU-CF: accuracy 0.78 at confidence 0.91). \textbf{Recalibrating
confidence does not close this gap}. Recalibrating forward in time, on
earlier months only, leaves post-boundary questions 0.19--0.23 overconfident.
At equal confidence above 0.6, post-boundary questions are 0.17--0.31 less
often right. Because $\pmax$ takes only 151 distinct values on this panel, all
seen on both sides, recalibration maps can be bounded directly. An exact
optimisation over contiguous partitions of the values, which relaxes
monotonicity, shows that no order-preserving map can bring the
period-conditional calibration error below 0.047 (a bound that need not be
attained). Searches over arbitrary
groupings improve this only modestly in sample (0.043--0.046) and are worse on
held-out months, where the cross-fitted contiguous-partition map leaves
post-boundary questions 0.18 overconfident; a map that also knows the period
closes both gaps (exploratory, on this news panel; \cref{app:everyf}). The
differences persist when the change point is re-selected inside a block
bootstrap and at every fixed cutoff from June 2024 to March 2025.
Distinguishing the two periods requires information beyond the confidence
score in these analyses.

\subsection{What targeted questions reveal, and what they do not}
\label{sec:followups}

\Cref{tab:gaps} is the paper's central result. For gaps in the case, a single
yes/no question asked on the same call gives a direct readout, sharper or
better oriented than confidence; for
gaps in the model's memory, the apparent signal does not survive controls.

\textbf{``Is the answer already settled?''} When a fact is fixed but unknown
and when an outcome awaits a fair random draw, a calibrated answer
distribution has the same ideal in both cases, so which kind of uncertainty is
at play is not identifiable from that distribution alone. Jev's actual
answer uncertainty does separate them (AUROC 0.96, higher for the fixed
facts), but only because Jev concentrates on the mode of a stated chance event,
itself a calibration failure (\cref{sec:format}): even in the uniform-chance
scenarios, where the ideal distributions are identical, its mean confidence is
0.76 on future draws against 0.49 on unknown facts (AUROC 0.89). Read as a
statement of knowledge, it points the wrong way. \texttt{settled} exceeds it by
0.04 [0.03, 0.06] overall and 0.11 [0.07, 0.16] in the uniform scenarios. The \texttt{settled} question separates them directly (mean $P(\text{yes})$ 0.87 vs.\ 0.13;
\cref{fig:scenarios}c), including traps that pit meaning against wording (a
past random draw is settled; a council decision next week is not) and made-up
events whose status must be inferred from tense (``was played last spring'' vs.\
``will be played next spring'') or from dates, provided today's date is given.

\textbf{``Is there enough information?''} This question gives a direct readout
of evidence completeness. On the same two-paragraph HotpotQA conditions it
separates complete from incomplete evidence with AUROC 0.95. Jev's confidence
also rises with complete evidence (AUROC 0.85, higher meaning complete), and
text length alone reaches 0.65; on the same items \texttt{enough} beats
confidence by $+0.10$ [0.09, 0.11], so the question adds a clearer signal
rather than one confidence lacks. Separately, misleading
context reduces accuracy without reducing confidence: with related but non-supporting
paragraphs, HotpotQA accuracy falls from 0.84 to 0.75 while confidence stays at
0.83; $P(\text{enough})$ is 0.13 for these and 0.86 with both supporting
paragraphs. All conditions with paragraphs show two of them, so paragraph
count does not distinguish them (text length is discussed below). When a deciding fact is deleted from a policy case, $P(\text{enough})$
falls from 0.83 to 0.10 and a ``cannot tell'' option receives 0.91.

\textbf{``Do you know the answer?''} This is the only one of the three about
the model's own knowledge. On the original contrasts it flags fabricated
entities (0.91, where $1-\pmax$ reaches 0.75; 0.88 against the same obscure
real people used below, where $1-\pmax$ reaches 0.78) and news past the boundary
(0.91, where confidence ranks post-boundary questions \emph{higher}), and it
beats the strongest self-evaluation baseline we ran, a second-look ``is the
proposed answer correct?'', on the same items (0.92 vs.\ 0.80 for made-up
entities, 0.91 vs.\ 0.62 for news); across time it falls
steadily from about 0.40 for 2020 news to about 0.10 for 2025--26 news
(\cref{fig:boundary}b).

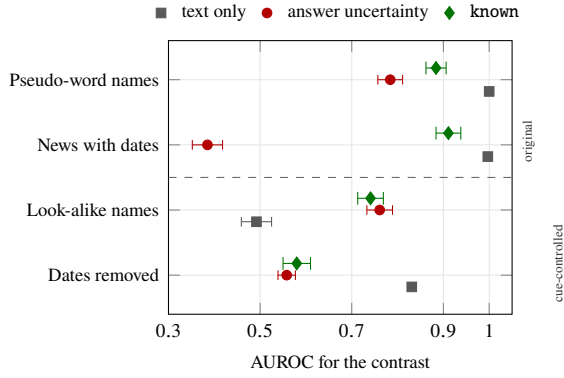
\begin{figure}[t]
\centering
\begin{tikzpicture}
\begin{axis}[
  width=0.74\columnwidth, height=5.2cm, scale only axis=false,
  xmin=0.3, xmax=1.05, ymin=0.4, ymax=4.6,
  xlabel={AUROC for the contrast}, xlabel style={font=\scriptsize},
  ytick={1,2,3,4},
  yticklabels={{Dates removed},{Look-alike names},{News with dates},{Pseudo-word names}},
  yticklabel style={font=\scriptsize, align=right},
  xticklabel style={font=\scriptsize},
  xtick={0.3,0.5,0.7,0.9,1.0},
  grid=major, grid style={gray!20},
  legend style={font=\scriptsize, at={(0.5,1.03)}, anchor=south, draw=none,
    legend columns=3, column sep=4pt},
  legend cell align=left,
  clip=false,
]
\draw[black!60, dashed] (axis cs:0.3,2.5) -- (axis cs:1.05,2.5);
\node[font=\tiny, anchor=south west, gray!40!black, rotate=90] at (axis cs:1.12,2.6) {original};
\node[font=\tiny, anchor=north west, gray!40!black, rotate=90] at (axis cs:1.12,0.45) {cue-controlled};
\addplot[only marks, mark=square*, mark size=1.8pt, gray!70!black,
  error bars/.cd, x dir=both, x explicit]
  coordinates {(0.831,0.82) +- (0.0,0) (0.492,1.82) +- (0.033,0) (0.997,2.82) +- (0.001,0) (1.000,3.82) +- (0.0,0)};
\addlegendentry{text only}
\addplot[only marks, mark=*, mark size=1.8pt, red!70!black,
  error bars/.cd, x dir=both, x explicit]
  coordinates {(0.558,1.0) +- (0.019,0) (0.761,2.0) +- (0.028,0) (0.385,3.0) +- (0.033,0) (0.784,4.0) +- (0.027,0)};
\addlegendentry{answer uncertainty}
\addplot[only marks, mark=diamond*, mark size=2.4pt, green!45!black,
  error bars/.cd, x dir=both, x explicit]
  coordinates {(0.580,1.18) +- (0.03,0) (0.741,2.18) +- (0.028,0) (0.911,3.18) +- (0.027,0) (0.884,4.18) +- (0.022,0)};
\addlegendentry{\texttt{known}}
\end{axis}
\end{tikzpicture}
\caption{Controlling for surface cues. On the original contrasts (upper
two rows) a text-only classifier that never queries Jev separates made-up from
real subjects and later from earlier news almost perfectly, and \texttt{known}
scores 0.88--0.91. With realistic names and with explicit temporal expressions
removed (lower two rows), the text classifier weakens and \texttt{known} falls
to the level of answer uncertainty ($1-\pmax$). Names rows use the same 555
obscure real people as controls (text-only for pseudo-word names: against the
least popular real subjects). Answer uncertainty below 0.5 for news with dates
means confidence is \emph{higher} after the boundary. Both news rows use the
same 3{,}360 questions (1{,}680 per period); the date-only text baseline is from
the full panel, and the masked-text baseline is month-grouped (0.83). 95\% intervals; for masked news the text classifier still uses
topic cues. Exploratory analyses, added after the main studies.}
\label{fig:cues}
\end{figure}

\textbf{Controlling for surface cues} (\cref{fig:cues}). These contrasts contain
cues that require no knowledge. Our fabricated names are pseudo-words: a character
n-gram classifier on the question text alone separates them from real
subjects perfectly (AUROC 1.00, also against the least popular real ones).
Likewise, 98.7\% of the news questions contain a year, and the date alone
separates the periods (0.997). Controlling these measured cues removes most
of \texttt{known}'s signal. Against the same 555 obscure real people and
relations, look-alike names built from real first names and surnames leave a
text-only classifier at chance (0.49), and \texttt{known} drops from 0.88
(pseudo-word names) to 0.74 [0.71, 0.77], similar to answer uncertainty (0.76
[0.73, 0.79]). With explicit temporal expressions removed from the news
questions, \texttt{known} falls to 0.58 [0.55, 0.61] (answer uncertainty 0.56
[0.54, 0.58]); this comparison uses only period membership, so it does not
depend on the rewritten questions' answer labels. A text classifier still
separates the masked periods at 0.83, so period information remains in the
text; Jev's familiarity answer uses little of it. On the same
items, the paired difference $\texttt{known}-(1-\pmax)$ is $-0.02$
[$-0.05$, 0.01] for realistic names, equivalent within $\pm0.05$ (90\% interval),
and $+0.02$ [$-0.01$, 0.06] with dates removed, where equivalence at that margin
is not shown but there is no demonstrated advantage; with dates present it is
$+0.53$. The equivalence margin, like these controls, was chosen after the
main studies. Grouping the text-classifier folds by subject or month
preserves the qualitative conclusions (1.00 and 0.50 for names; 0.83 for
masked news, against 0.86 ungrouped). \texttt{known} does respond to
information in the input: when one sentence states the answer about a made-up
subject, P(\texttt{known}) is 0.81 with an answer-bearing sentence, 0.20
without a sentence and 0.05 with an unrelated sentence. Within popularity levels or
news months it ranks right against wrong answers only weakly (0.48--0.66). On
these tests it therefore tracks whether an answer is available in the input or
looks familiar; we find no robust additional signal about whether Jev holds
the knowledge, although fabricated-versus-real status and period are only
proxies for that. By contrast, \texttt{enough} is not explained by length.
Among HotpotQA conditions with two paragraphs, complete evidence is the
\emph{shortest}, so length separates them with AUROC 0.65 in the ``shorter means
complete'' direction; comparing only items within the same 10-word length
stratum (a size-weighted mean of within-stratum AUROCs), \texttt{enough} keeps 0.95
while length falls to 0.48, and adding \texttt{enough} to a gradient-boosted
model of length features (question and paragraph lengths) raises AUROC from
0.60 to 0.94. For pairs from the same question within 10 words of each other,
\texttt{enough} ranks the complete one higher 96\% of the time.

\textbf{Scope.} Each question targets a specific gap. \texttt{settled} and
\texttt{enough} judge the input; \texttt{known} tracks surface familiarity and
supplied evidence, and is a weak detector of individual errors on ordinary
questions (TriviaQA AUROC 0.76 vs.\ 0.96 for $\pmax$). \texttt{enough} detects
incomplete evidence, not wrong answers: Jev often answers correctly without
the paragraphs, so accepting answers by \texttt{enough} is 2--6 points less
accurate than accepting by confidence at the same coverage (thresholds set on
held-out questions). The
rankings are robust to fourteen rewordings, but absolute levels vary, so
thresholds must be set per wording. These AUROCs measure discrimination of
curated contrasts; the answers are ranking scores, not validated probabilities
of an epistemic state. Ambiguity is flagged only weakly
(``single clear answer?'' AUROC 0.66).

\subsection{Putting the questions to use}
\label{sec:use}

\begin{figure*}[t]
\centering
\includegraphics[width=\textwidth]{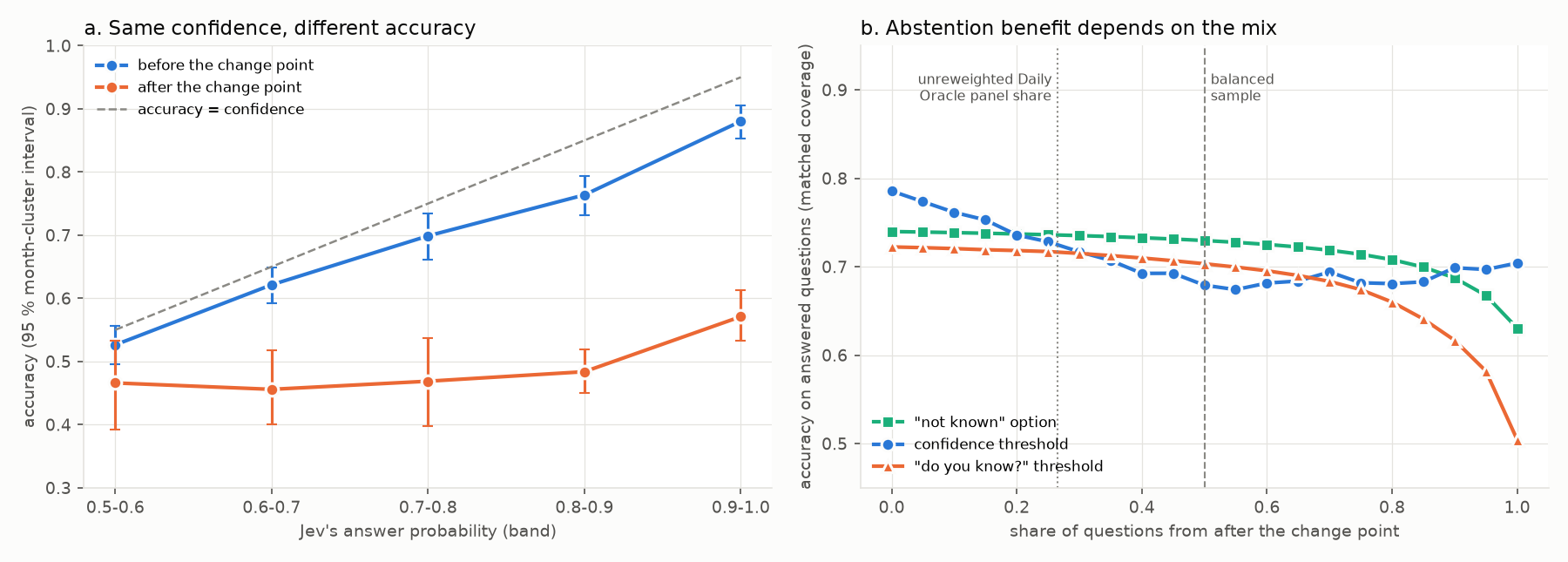}
\caption{Daily Oracle yes/no questions around the change point (2024-11).
(a) Accuracy at equal confidence (95\% month-cluster intervals):
lower and nearly flat after the change point, so a common map of
confidence cannot match accuracy at each value on both sides. (b) Accuracy at matched coverage as the
share of post-change questions varies: a ``not known'' option beats a
confidence threshold when such questions are common and beats thresholding
``do you know?'' at every mix. The dotted line marks the share in our date
panel, not a deployment prevalence.}
\label{fig:recal}
\end{figure*}

\textbf{Combining.} Adding $P(\text{known})$ to confidence raises
error-detection AUROC on dated news from 0.61 to 0.66, and a recalibration map
that uses both cuts the post-boundary gap from 0.30 (raw) and 0.24 (confidence
alone) to 0.16; a map that knows the period closes it
(\cref{app:everyf}). On ordinary questions $P(\text{known})$ adds nothing,
because there confidence already works.

\textbf{A ``not known'' option} puts the \texttt{known} idea into the answer
itself. Jev chooses it for 93\% of post-boundary news questions, so it answers
only about 7\% of them, and for 36\% of earlier ones; confident errors nearly
vanish among the questions it answers, a count that depends on the format,
since the third option changes the probability scale. At matched coverage it
beats answering by confidence when half the questions lie past the boundary
($+0.05$ accuracy), is inconclusive at the 27\% share of our date panel
($+0.01$ [$-0.00$, 0.04]), and is worse when such questions are rare ($-0.03$ at
5\%; \cref{fig:recal}b). It routes boundary questions rather than ranking
errors, and it beats thresholding \texttt{known} at every mix. These questions
contained their dates, which likely help the option as they help
\texttt{known}. A cheap rule gets close: skipping the questions whose own
latest date is most recent, at the option's coverage and using Jev's ordinary
answers, trails the option by only 0.01--0.02 at every mix (at the panel's
share, $+0.02$ [0.00, 0.04] in the option's favour; coverage and error rates
in \cref{tab:coverage}).

\textbf{Stating that context may be irrelevant} restores HotpotQA accuracy with
misleading paragraphs (0.75 $\to$ 0.84) and calibration (SmoothECE 0.02). A
second-look check (``is the proposed answer correct?'') reduces estimated
pooled calibration error on the balanced news set of 1{,}680 questions from each
period (SmoothECE 0.18 $\to$ 0.03), which does not by itself establish
calibration within each period, but makes easy sets
underconfident. Giving today's date alone barely helps.

\subsection{For stated chances, ask one question per outcome}
\label{sec:format}

\begin{figure*}[t]
\centering
\includegraphics[width=\textwidth]{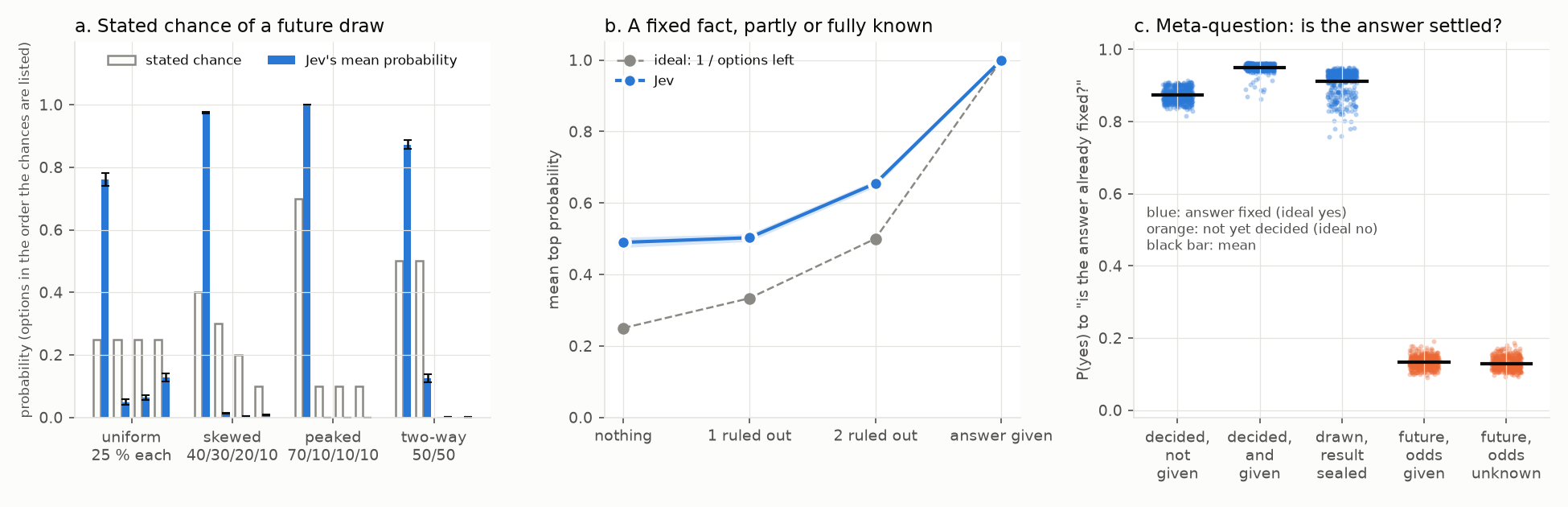}
\caption{Knowledge$\times$chance scenarios (480 generated questions). (a) Stated
chances of a future draw vs.\ Jev's Choice probabilities: nearly all mass on
the likeliest outcome, ties broken towards the outcome mentioned first. (b) A
fixed fact, progressively revealed: right direction, too confident
throughout. (c) The \texttt{settled} question separates fixed-but-unknown
answers from future draws perfectly.}
\label{fig:scenarios}
\end{figure*}

\begin{table}[t]
\centering
\small
\setlength{\tabcolsep}{2.6pt}
\caption{Mean reading by stated chance (480 scenarios).}
\label{tab:format}
\begin{tabular}{@{}l rrrrrr r@{}}
\toprule
Stated & 0 & .1 & .2 & .3 & .4 & .7 & MAE \\
\midrule
Yes/no per outcome & .01 & .09 & .16 & .25 & .36 & .68 & .030 \\
Score per outcome & .05 & .13 & .22 & .34 & .45 & .74 & .028 \\
Instructed Choice & .00 & .00 & .00 & .01 & .99 & 1.0 & .230 \\
\bottomrule
\end{tabular}
\end{table}

The same lesson holds for aleatoric uncertainty. Told that a future draw yields
outcomes with chances 40, 30, 20 and 10\%, Jev's Choice puts 0.98 on the 40\%
outcome (\cref{fig:scenarios}a), and instructing it to report chances changes
nothing. Asked one outcome at a time, the yes/no probability is within 0.030 of
the stated chance on average (\cref{tab:format}); the readings sum to 0.89, and
rescaling them to one lowers the error to 0.012. Our Choice prompts ask for
the realised outcome (e.g.\ ``What is the colour of the guild's banner at the
coming festival?'', instruction ``Answer the question using the facts
given''), and the vendor describes a Choice as returning the full
probability distribution over the options, with probabilities optimised
against outcomes \citep{typesafe2026choice,typesafe2026systemone}. Read that
way, 0.98 on a 40\% outcome is badly miscalibrated. Rephrasing does not change
it: asking which outcome ``will come up'', which is ``most likely'', or which
one ``you win if you pick'' gives the same concentration on the mode (loaded
die: 1.00 on its 50\% face; sum of two dice: 0.98--1.00 on 7). Only asking for
``the probability of each'' moves part-way towards the stated odds (loaded die
0.66; two dice 0.54 on 7, stated 0.17). Under no tested wording did a Choice
distribution behave as an outcome distribution, and ties fail too: 0.80 on ``one'' for a fair die,
following labels rather than display positions, which survives averaging
over option orders \citep{zheng2024large}. For outcome probabilities, ask one
yes/no question per outcome.

\section{Discussion}
\label{sec:discussion}

Jev's confidence and the follow-up questions play different roles.
Confidence is the probability Jev assigns to its chosen option: calibrated
on the familiar closed-choice benchmarks we tested, where it supports
abstention (e.g.\ at $\pmax\ge0.9$), although we did not evaluate deployment
utility under other prevalences, costs or shifts. It is not a readout of what the model does not know---when
nothing decides the answer, Jev still names one, chosen by salience, and past
its knowledge boundary its confidence goes up, with substantial sensitivity
to the dates in the questions. Targeted questions provide more direct or
sharper readouts of particular case properties than answer confidence:
whether an outcome is settled and whether the evidence supplied suffices. Asking Jev whether it
knows the answer looked like a readout of its own knowledge, but controls show
that it largely tracks unfamiliar-looking names, explicit dates and supplied
evidence; once the measured cues are removed it shows no demonstrated
advantage over confidence. For Jev, gaps in the case are therefore better
\emph{queried} than read from a single distribution, whereas the familiarity
question provides no evidence of memory-gap detection beyond confidence once
the measured cues are controlled; our labels of such gaps are proxies. This is a case study of one model;
whether the same holds for other decision models is untested. We do not claim
these questions recover a general epistemic state; each catches the gap it
asks about.

\paragraph{Practical guidance.} On familiar, in-scope closed-choice questions
like those tested, Jev's confidence can be used for abstention; a single call's aggregate calibration is close
to that of three. Add \texttt{settled},
\texttt{enough} where outcomes may be undetermined or context incomplete; do
not rely on \texttt{known} to detect gaps in the model's memory, and set thresholds per wording on held-out
data. Evaluate a ``not known'' option at the expected input mix and error
costs; in our news comparisons it improved matched-coverage accuracy only when
later-period questions were common: it answers few such questions (about 7\% past the boundary),
is inconclusive at our panel's mix and costs accuracy when they are rare, and a
simple rule that skips questions with the most recent dates comes within
0.01--0.02 of it; give the current date when asking whether an outcome is
settled, and state that retrieved context may be irrelevant. To recover explicitly stated outcome probabilities, ask one
yes/no question per outcome and rescale the readings to sum to one. This is
validated against the stated odds, including expected proper scores under the
stated process (\cref{app:scores}); for real-world outcomes it is untested.

\section{Conclusion}

On the familiar closed-choice benchmarks we tested, Jev's confidence tracks
how likely an answer is to be right, but it does not show what the model does
not know: with nothing to go on
it backs a salient guess, and past its knowledge boundary it grows more
confident in a way that confidence-only recalibration does not repair.
Targeted questions on the same call---is the outcome settled, is there enough
information---discriminate gaps in the case on the tested contrasts, while
asking whether it knows the answer mostly reads surface cues. Gaps in the case
can be asked about; gaps in the model's knowledge remain hard to see from
outside, and auditing them requires
controls that remove the cues a model can read instead.

\section*{Code availability}
The code for the experiments and analyses is available at
\url{https://github.com/Syntheme/beyond-answer-confidence}.

\section*{Limitations}
\label{sec:limitations}

\textbf{Scope.} This is a case study of one version of one closed model; no
other system was run through the full protocol, so we make no claim about
decision models or LLMs in general. \textbf{Surface cues.} The follow-up
questions are evaluated on curated contrasts. For \texttt{known}, text-only
baselines, realistic names and date removal show that surface cues carry most
of its signal; a text classifier at chance does not certify that no cue
remains. \texttt{settled} is by design a judgement about the input, and
\texttt{enough} was checked against text length and length features only.
\textbf{Date removal.} Removing temporal expressions can change a question's
proposition (e.g.\ by dropping a deadline) or remove which event is meant,
so we report response changes, not validated accuracy; the provisional labels
of 240 pairs come from a single annotator (an AI assistant following a written
protocol). \textbf{Observational boundary.} The knowledge boundary is an
accuracy change point in dated news, not a verified training cutoff, and
topic or style may change with it. \textbf{Forced choice.} Most benchmarks
were converted to multiple choice. Fabricated-entity questions have no valid
answer (a false premise), so their 0.25 is a uniform forced-choice reference;
made-up future events have a defined predictive target that is symmetric by
construction; and fixed but unknown generated facts have a hidden answer with
a uniform prior. \textbf{Probability semantics.} The vendor does not
specify whether a Choice models the realised outcome or the best answer.
\textbf{Scores, not probabilities.} The follow-up answers are validated as
rankings, not as calibrated probabilities of an epistemic state.
\textbf{Decision value.} Benefits of the ``not known'' option depend on how
often boundary questions occur. Some answer-filter comparisons use thresholds
chosen on held-out questions, but cost-based utility and a complete deployment
routing policy were not evaluated. \textbf{Exploratory analyses.} The surface-cue
controls, recalibration bounds and robustness analyses were added after the
main studies and reuse the same dated-news panel; the temporal findings have
not been confirmed on a fresh panel.

{\footnotesize
\setlength{\bibsep}{1pt plus 0.3ex}
\bibliographystyle{plainnat}
\bibliography{refs}
}

\clearpage
\onecolumn
\appendix

\section{Example request and response}
\label{app:example}

One call from the knowledge$\times$chance scenarios (option ``blue'' ruled out,
so the ideal Choice answer is 1/3 on each remaining option); the response is
shown as returned, with usage counts omitted.

\begin{small}
\begin{verbatim}
state: {"facts": [
  "Voskin is mentioned in 31 old records.",
  "The name Voskin comes from an old word meaning 'bridge'.",
  "The colour of the Voskin guild's banner at the coming
   festival has already been decided.",
  "The colour of the Voskin guild's banner at the coming
   festival is not blue."],
 "question": "What is the colour of the Voskin guild's banner
   at the coming festival?"}
questions: {
 "answer":   {"type": "choice",
              "instructions": "Answer the question using the facts given.",
              "criteria": {"red": null, "green": null,
                           "blue": null, "black": null}},
 "settled":  {"type": "noul",
              "instructions": "Is the answer to the question already fixed
                at this moment, so that someone with more information than
                is given here could know it for certain?"}}

response: {"model": "jev-1.13.0", "answers": {
 "answer":  {"type": "choice", "choice": "red", "confidence": 0.46,
             "probabilities": {"red": 0.60, "green": 0.23,
                               "black": 0.15, "blue": 0.02}},
 "settled": {"type": "noul", "noul": 0.89}}}
\end{verbatim}
\end{small}

The request also carried a second follow-up question (``do the facts
determine the answer with certainty?'', $P(\text{yes})=0.04$), omitted here.

\section{Statistical methods and robustness}
\label{app:stats}

\paragraph{Intervals.} Intervals are 95\% item-clustered bootstrap
intervals with 2{,}000 resamples; dated news uses months as clusters, and
time-dependent analyses use period-stratified 6-month moving blocks. For
paired AUROC comparisons, resampling is by original question (HotpotQA) or
generated scenario, retaining all associated conditions and all compared
scores. We report effect sizes and intervals rather than pass/fail tests.

\subsection{A single call}
\label{app:single}
Over 500 random draws of one of the three cached calls per item, single-call
SmoothECE stays within 0.010 of the three-call value (95\% range) and accuracy
within 0.01 in all 36 calibration sets. The point estimate falls on the same
side of 0.05 as the three-call value in every draw except SimpleQA Verified
(three-call 0.051; below 0.05 in 11\% of draws).

\begin{table}[h]
\centering\small
\caption{SmoothECE of the mean of three calls and of one random call per item (500 draws).}
\begin{tabular}{@{}l r r r r@{}}
\toprule
Set & $n$ & mean of 3 & single, 2.5--97.5\% & estimate $<0.05$ \\
\midrule
Banking77, 8 examples & 770 & 0.037 & 0.031--0.040 & 100\% \\
CLINC150, 1 example & 1{,}500 & 0.025 & 0.021--0.026 & 100\% \\
PopQA & 14{,}267 & 0.021 & 0.021--0.025 & 100\% \\
Daily Oracle, after change point & 1{,}680 & 0.305 & 0.304--0.310 & 0\% \\
TriviaQA & 9{,}960 & 0.031 & 0.030--0.032 & 100\% \\
SimpleQA Verified & 1{,}000 & 0.051 & 0.048--0.061 & 11\% \\
MMLU-Redux & 5{,}330 & 0.016 & 0.015--0.018 & 100\% \\
MMLU-CF & 10{,}000 & 0.145 & 0.144--0.147 & 0\% \\
\bottomrule
\end{tabular}
\end{table}

\subsection{Coverage of the SmoothECE intervals}
Confidences were drawn from four real Jev confidence pools at their real
sizes, with correctness Bernoulli$(\min(1,\max(0,p-\delta)))$ and $\delta$ set
so that the population SmoothECE is 0, 0.03, 0.05 or 0.07 (400 datasets per
setting). At a true value of exactly 0.05, the one-sided 95\% upper bound falls
below 0.05 in 1--6\% of datasets for the plain percentile bootstrap (nominal
5\%), 2--9\% for a bias-recentred bootstrap, and 13--28\% for the basic
bootstrap and half-sample subsampling. We therefore report percentile bounds.
The 0.05 threshold is used only descriptively. The
estimator is biased upward by 0.010--0.025 for perfectly calibrated data, so
small-sample values of 0.02--0.03 largely reflect that floor. Only one shape of
overconfidence was simulated.

\subsection{The knowledge change point}
The change point is the month that best splits monthly yes/no accuracy into two
means. Re-selecting it in 2{,}000 resamples, a drop of at least 0.05 is always
found (2024-11 in 49\% of resamples, mostly 2024-01 to 2024-10 otherwise);
the post-change gap between confidence and accuracy is 0.21--0.33 and the rise
in confidence $+0.02$ to $+0.07$. Every fixed cutoff from 2024-06 to 2025-03
gives a gap of 0.26--0.32. With flexible time trends (natural splines, HAC
intervals), $P(\text{known})$ shows no step at the change point
($-0.014$ [$-0.033$, 0.005]); it declines steadily.

\subsection{Recalibration at matched confidence values}
\label{app:everyf}
On the Daily Oracle yes/no questions, $\pmax$ takes 151 distinct values, all
observed in both periods. A recalibration map assigns an output level to each
value; for a level $c$ shared by a group of values, its period-conditional
error is $n_{\text{pre}}|c-a_{\text{pre}}| + n_{\text{post}}|c-a_{\text{post}}|
\ge \min(n_{\text{pre}},n_{\text{post}})\,|a_{\text{pre}}-a_{\text{post}}|$,
computed on the group's pooled accuracies $a$. Merging values can cancel
differences between periods, so we evaluate three classes of map
(\cref{tab:recal}). The value-wise bound (each value its own level) is 0.066;
over all contiguous partitions of the values, solved exactly by dynamic
programming without requiring the levels to increase, the minimum is 0.047,
which is therefore a lower bound for every monotone map; random sets of 21
months give 0.005 (descriptive $p=0.005$). A local search over arbitrary
groupings finds 0.043--0.046. Cross-fitted on held-out months, every class
leaves a large error, and the cross-fitted contiguous-partition map leaves
post-change questions 0.18 overconfident. Recalibrating forward in time on earlier months
only leaves $+0.19$ after the change point with an expanding window and
$+0.23$ with a map frozen at the change point (+0.17 to +0.31 by month). A
period-aware isotonic map closes both gaps, and a logistic model with period
terms gives $\chi^2_2=281$. The confidence-band intervals depend on the
procedure: with the change point fixed at 2024-11 and months resampled within
each period (\cref{tab:bands}), the lowest band's interval includes zero
($-0.014$ to 0.139); re-selecting the change point in every period-stratified
6-month-block resample, all five exclude zero (lowest band 0.015 to 0.123).

\paragraph{Optimisation.} Let distinct confidence values be sorted, and let a
map group them and assign each group one output level $c$. For a group with
$n_{\text{pre}}, n_{\text{post}}$ items and pooled accuracies $a_{\text{pre}},
a_{\text{post}}$, the smallest achievable error $n_{\text{pre}}|c-a_{\text{pre}}|
+ n_{\text{post}}|c-a_{\text{post}}|$ over $c\in[0,1]$ is $\min(n_{\text{pre}},
n_{\text{post}})|a_{\text{pre}}-a_{\text{post}}|$; the total is summed over groups and
divided by all 6{,}320 questions. The interval result minimises this total over
every partition of the sorted values into contiguous groups, by dynamic
programming ($E_j=\min_{i<j}[E_i+\text{cost}(i,j)]$), without requiring the
levels to increase. Since every monotone map's level sets are contiguous, this
relaxation is a lower bound on the error of every monotone map. The
comparison distribution repeats the optimisation after drawing 21 of the 79
months at random as ``after'' (200 draws). It treats months as exchangeable,
ignoring drift, serial dependence and the selection of the change point, so
its $p$-value is descriptive; a temporally structured null (contiguous placebo
boundaries or moving blocks with the boundary re-selected) is not reported; cross-fitting splits months within each period in
half (50 random splits, used in both directions). A fitted map is applied as
follows: the partition is chosen on the training months using their period
labels (the objective is period-conditional), each group's output level is
the training accuracy of its larger period (the minimiser of the group cost;
ties go to the earlier period), and a held-out value whose group has no
training items receives the overall training accuracy; held-out error is
then scored on the fitted groups. This is a partition-conditioned
discrepancy: two groups that receive the same output level are scored
separately, so it need not equal calibration error after equal output levels
are pooled. The map itself takes only confidence as input. The cross-fitted error
averages over both periods, while the post-change gap is the mean signed
overconfidence after the change point, so the two numbers measure different
things.
These are post-hoc results for this news panel.

\begin{table}[h]
\centering\small
\caption{Period-conditional calibration error of confidence-only recalibration
maps on dated news.}
\label{tab:recal}
\begin{tabular}{@{}>{\raggedright\arraybackslash}p{0.36\linewidth} r >{\raggedright\arraybackslash}p{0.2\linewidth} >{\raggedright\arraybackslash}p{0.2\linewidth}@{}}
\toprule
Maps allowed & In-sample & Random months (descriptive), mean / 95th pct & Cross-fitted mean (range) \\
\midrule
Value-wise (each value its own level) & 0.066 & 0.034 / 0.038 & 0.170 (0.147--0.191) \\
Contiguous partitions (exact; lower bound for monotone maps) & 0.047 & 0.005 / 0.013 & 0.057 (0.046--0.074) \\
Arbitrary groupings, 2--20 levels (search) & 0.043--0.046 & --- & 0.093--0.099 \\
\bottomrule
\end{tabular}
\end{table}

\begin{table}[h]
\centering\small
\caption{Accuracy by answer-probability band before and after the change
point (month-cluster intervals).}
\label{tab:bands}
\begin{tabular}{@{}l r r r r r@{}}
\toprule
$\pmax$ band & $n$ before & acc.\ before & $n$ after & acc.\ after & Difference [95\% CI] \\
\midrule
0.5--0.6 & 988 & 0.526 & 206 & 0.466 & 0.060 [$-0.014$, 0.139] \\
0.6--0.7 & 943 & 0.621 & 204 & 0.456 & 0.166 [0.100, 0.228] \\
0.7--0.8 & 777 & 0.699 & 224 & 0.469 & 0.230 [0.152, 0.306] \\
0.8--0.9 & 821 & 0.764 & 370 & 0.484 & 0.280 [0.236, 0.327] \\
0.9--1.0 & 1{,}111 & 0.880 & 676 & 0.571 & 0.309 [0.260, 0.354] \\
\bottomrule
\end{tabular}
\end{table}

\section{Additional results}
\label{app:more}

\subsection{Surface-cue controls for the follow-up questions}
\label{app:cues}

Text-only baselines are character 2--5-gram TF-IDF features with logistic
regression, scored out of fold (\cref{tab:cues}). Look-alike made-up names
join the first name of one obscure PopQA person to the surname of another (100
per person relation; the question template and options are as for the
pseudo-word items). Date masking removes years, month names, weekdays, day
numbers, seasons and phrases such as ``by the end of'' from the news questions
(``Will the Fed cut rates by the end of March 2024?'' $\to$ ``Will the Fed cut
rates?''); 6{,}274 of 6{,}320 questions change. With dates removed,
post-change ``no'' answers fall from 92\% to 67\%; accuracy is reported only on
answer-safe subsets (\cref{app:dates}). When one sentence
states the answer about a made-up subject, P(\texttt{known}) is 0.81, against
0.05 with an unrelated sentence of the same form and 0.20 with none, and Jev
picks the stated option every time. Asked to verify both a proposed ``yes''
and a proposed ``no'' for every news item, Jev gives P(correct) 0.26 and 0.56
after the change point (0.40 and 0.54 before); with the negative wording
(``is the proposed answer wrong?'') the implied answers agree with the positive
wording on 95--96\% of items, imply ``yes'' for 10\% of post-change questions
(true share 48\%), and accept both proposals for 0.1\%. The true share of ``no''
is 51.2\% before and 51.5\% after; the category mix differs, but accuracy falls within
every news category, so shifts between broad categories alone do not explain
the drop.

\begin{table}[h]
\centering\small
\caption{AUROC for the \texttt{known} contrasts with and without surface cues.
Each row uses one set of items (made-up / real, or after / before).}
\label{tab:cues}
\begin{tabular}{@{}l r r r r@{}}
\toprule
Contrast (items) & Text only & \texttt{known} & $1-\pmax$ & Second look \\
\midrule
Pseudo-word made-up vs.\ all real (1{,}600 / 14{,}267) & 1.000 & 0.911 & 0.746 & --- \\
Pseudo-word made-up vs.\ real, second-look subset (1{,}600 / 2{,}000) & --- & 0.916 & 0.747 & 0.80 \\
Pseudo-word made-up vs.\ least popular real & 1.000 & 0.833 & 0.685 & --- \\
Pseudo-word made-up vs.\ obscure real persons (600 / 555) & --- & 0.884 & 0.784 & --- \\
Look-alike made-up vs.\ obscure real persons (600 / 555) & 0.492 & 0.741 & 0.761 & --- \\
News after vs.\ before, full panel (1{,}680 / 4{,}640) & 0.997 & 0.913 & 0.389 & --- \\
News after vs.\ before, subset (1{,}680 / 1{,}680) & --- & 0.911 & 0.385 & 0.62 \\
News, dates removed, same subset & 0.831 & 0.580 & 0.558 & --- \\
\bottomrule
\end{tabular}
\end{table}

Paired differences on the same items ($\texttt{known}$ minus answer uncertainty $1-\pmax$,
bootstrap over items or months): realistic names $-0.020$ [$-0.054$, 0.013]
(90\% interval $-0.048$ to 0.007, inside $\pm0.05$); news with dates removed
$+0.023$ [$-0.012$, 0.056] (90\% interval $-0.007$ to 0.051); news with dates
$+0.526$ [0.470, 0.578]. With text-classifier folds grouped by subject or month,
the baselines are 1.000 (pseudo-word names), 0.498 (realistic names) and 0.831
(news with dates removed; 0.862 ungrouped).

\subsection{Date removal and answer labels}
\label{app:dates}

Temporal expressions (years, month names, weekdays, day numbers, seasons and
phrases such as ``by the end of'') were removed together with their
prepositions. A random sample of 240 (original, masked) pairs, 120 per period,
was labelled against a written protocol by a single annotator, an AI assistant;
a check by a second, human annotator is pending. Labels: \emph{answer
preserved} (the same event, and the original label holds under ``has this
happened by now?''; 83 pairs, all but one with label ``yes''), \emph{answer
uncertain} (a deadline or window removed from a ``no'' question; 77) and
\emph{identity lost or ill-posed} (the date identified which occurrence, the
question asked about a state at a time, or the edit broke it; 80). Dropping a
deadline alone cannot turn a ``yes'' into a ``no'', but 30 of the 112 ``yes''
pairs lost event identity, so ``yes'' questions are not all answer-safe. We
therefore report only response changes: among all 1{,}648 ``yes'' questions,
removing dates cut post-change ``no'' answers from 92\% to 57\% and pre-change
ones from 41\% to 37\%. On the pairs judged answer-preserving: 92\% to 65\%
after ($n=37$), a drop of 0.27 [0.14, 0.43], with ten answers changing from
``no'' to ``yes'' and none the other way; 39\% to 35\% before ($n=46$), a drop of
0.04 [$-0.04$, 0.13]. No accuracy on rewritten questions is used as label-valid. Setting today's date to just
after each event (the end of the month after the latest date the question
mentions) leaves post-change accuracy at 0.514 and ``no'' answers at 91\%.

\begin{table}[h]
\centering\small
\caption{The ``not known'' option on dated news: coverage, accuracy on
answered questions and confident errors (wrong with $\pmax\ge0.9$).}
\label{tab:coverage}
\begin{tabular}{@{}l l r r r r@{}}
\toprule
Condition & Period & Coverage & Accuracy & Conf.\ errors (all) & Conf.\ errors (answered) \\
\midrule
Base & before & 100\% & 0.697 & 2.4\% & 2.4\% \\
Base & after & 100\% & 0.511 & 17.3\% & 17.3\% \\
Today's date given & after & 100\% & 0.511 & 12.7\% & 12.7\% \\
``Not known'' allowed & before & 64.3\% & 0.740 & 0.3\% & 0.5\% \\
``Not known'' allowed & after & 6.6\% & 0.631 & 0.1\% & 0.9\% \\
Date + ``not known'' & after & 28.5\% & 0.491 & 0.1\% & 0.2\% \\
\bottomrule
\end{tabular}
\end{table}

\subsection{Proper scoring rules}
\label{app:scores}

\begin{table}[h]
\centering\small
\caption{Multiclass Brier score and log loss of the three-call mean
distribution, with the values for a uniform answer over $K$ options. Log loss
uses $-\log\max(p,10^{-4})$ without renormalisation, because Jev returns exact
zeros; the floor therefore bounds, rather than measures, the loss on those
items. With floors of $10^{-2}$ to $10^{-6}$, MMLU-CF log loss ranges from 0.80
to 1.34 (5.8\% of answers have probability zero on the correct option);
TriviaQA (0.130--0.135) and dated news (0.709--0.711) barely change.}
\label{tab:scores}
\begin{tabular}{@{}l r r r r r r r@{}}
\toprule
Set & $K$ & $n$ & Accuracy & Brier & Uniform & Log loss & Uniform \\
\midrule
TriviaQA & 4 & 9{,}960 & 0.955 & 0.067 & 0.750 & 0.132 & 1.386 \\
PopQA (real subjects) & 4 & 14{,}267 & 0.708 & 0.378 & 0.750 & 0.719 & 1.386 \\
Daily Oracle yes/no & 2 & 6{,}320 & 0.651 & 0.474 & 0.500 & 0.710 & 0.693 \\
HotpotQA, non-supporting paragraphs & 2 & 951 & 0.751 & 0.342 & 0.500 & 0.542 & 0.693 \\
HotpotQA, both supporting paragraphs & 2 & 951 & 0.958 & 0.071 & 0.500 & 0.193 & 0.693 \\
MMLU-CF & 4 & 10{,}000 & 0.776 & 0.377 & 0.750 & 1.075 & 1.386 \\
\bottomrule
\end{tabular}
\end{table}

For chance tasks, expected scores when outcomes follow the stated odds (ideal
in parentheses): future draw with stated percentages, Brier 0.919 (0.608), log
loss 2.36 (1.08); fair die, 1.321 (0.833) and 2.98 (1.79); loaded die, 1.000
(0.700) and 4.61 (1.50).

\begin{figure}[h]
\centering
\includegraphics[width=0.8\linewidth]{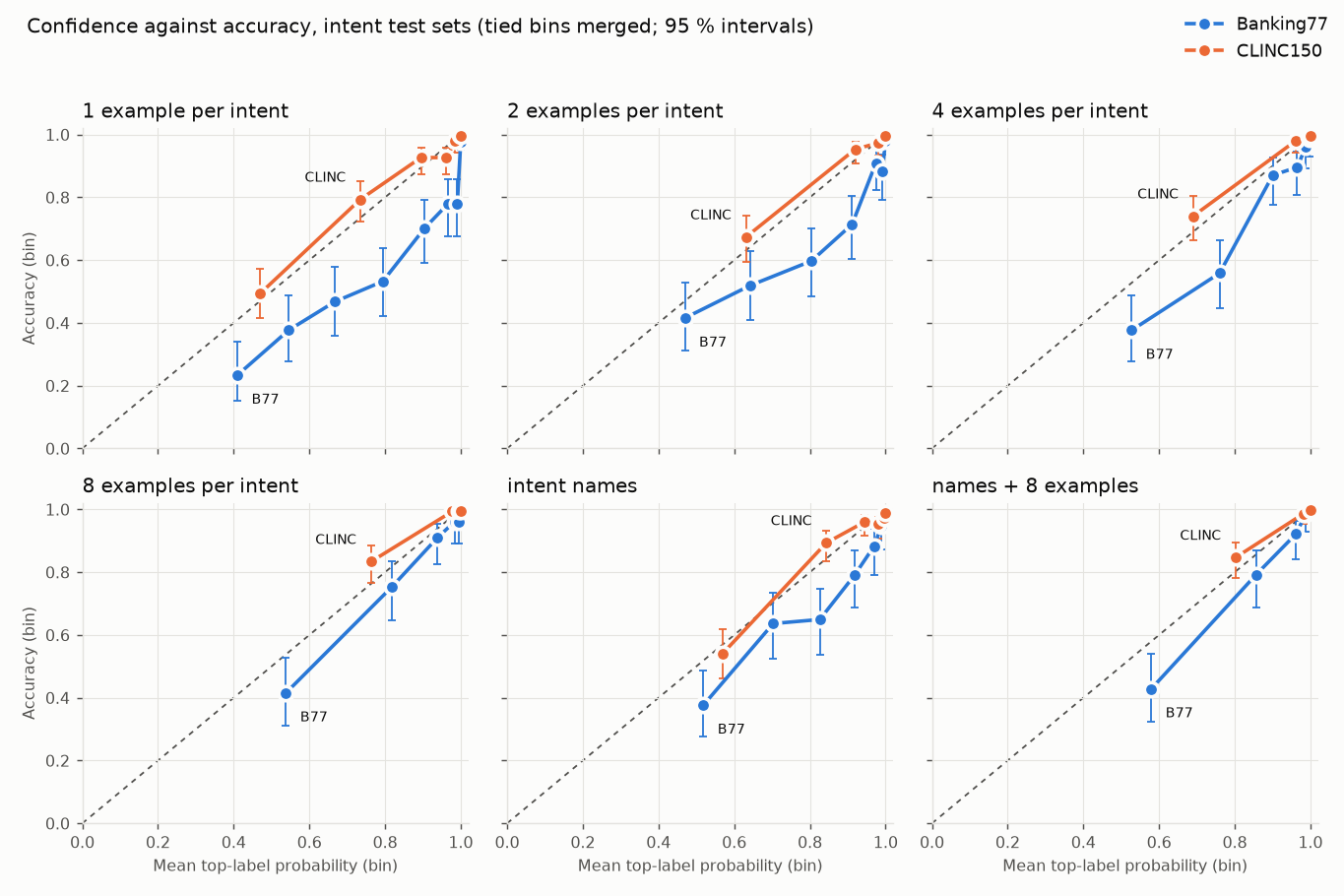}
\caption{Intent dose--response reliability by what each intent code carries
(about ten equal-count bins, with bins that would split identical confidence
values merged; 95\% intervals).}
\end{figure}

\begin{table}[h]
\centering\small
\caption{Intent dose--response, all conditions (mean of three calls).}
\begin{tabular}{@{}l rrrr rrrr@{}}
\toprule
& \multicolumn{4}{c}{Banking77} & \multicolumn{4}{c}{CLINC150} \\
\cmidrule(lr){2-5}\cmidrule(lr){6-9}
Code carries & acc & $\pmax$ & $\Hn$ & $\sece$ & acc & $\pmax$ & $\Hn$ & $\sece$ \\
\midrule
nothing & 0.010 & 0.320 & 0.774 & 0.310 & 0.008 & 0.363 & 0.592 & 0.355 \\
1 example & 0.681 & 0.828 & 0.109 & 0.170 & 0.911 & 0.904 & 0.057 & 0.024 \\
2 examples & 0.796 & 0.879 & 0.074 & 0.099 & 0.958 & 0.953 & 0.028 & 0.019 \\
4 examples & 0.861 & 0.914 & 0.056 & 0.066 & 0.969 & 0.964 & 0.021 & 0.015 \\
8 examples & 0.899 & 0.927 & 0.046 & 0.037 & 0.979 & 0.974 & 0.016 & 0.016 \\
name & 0.821 & 0.893 & 0.069 & 0.084 & 0.926 & 0.933 & 0.037 & 0.025 \\
name + 8 & 0.909 & 0.939 & 0.040 & 0.039 & 0.981 & 0.978 & 0.013 & 0.015 \\
irrelevant text & 0.019 & 0.312 & 0.559 & 0.293 & 0.007 & 0.226 & 0.579 & 0.219 \\
swapped examples & 0.001 & 0.932 & 0.044 & 0.904 & 0.001 & 0.974 & 0.016 & 0.957 \\
\bottomrule
\end{tabular}
\end{table}

\paragraph{Comparators.} GLiNER2.5-Decide, given the same items and codes, is
almost exactly uniform with no information ($\pmax$ 0.017 and 0.010), shows no
first-code preference, and reduces entropy with knowledge, but is
underconfident once informed (CLINC150 with names: $\pmax$ 0.34 at accuracy
0.66). A five-member MiniLM deep ensemble \citep{lakshminarayanan2017simple}
trained on the same few examples shows mutual information falling with more
data and stays cautious with one example ($\pmax$ 0.43--0.54) where Jev, more
accurate, is confident (0.83--0.90).

\begin{figure}[h]
\centering
\includegraphics[width=\linewidth]{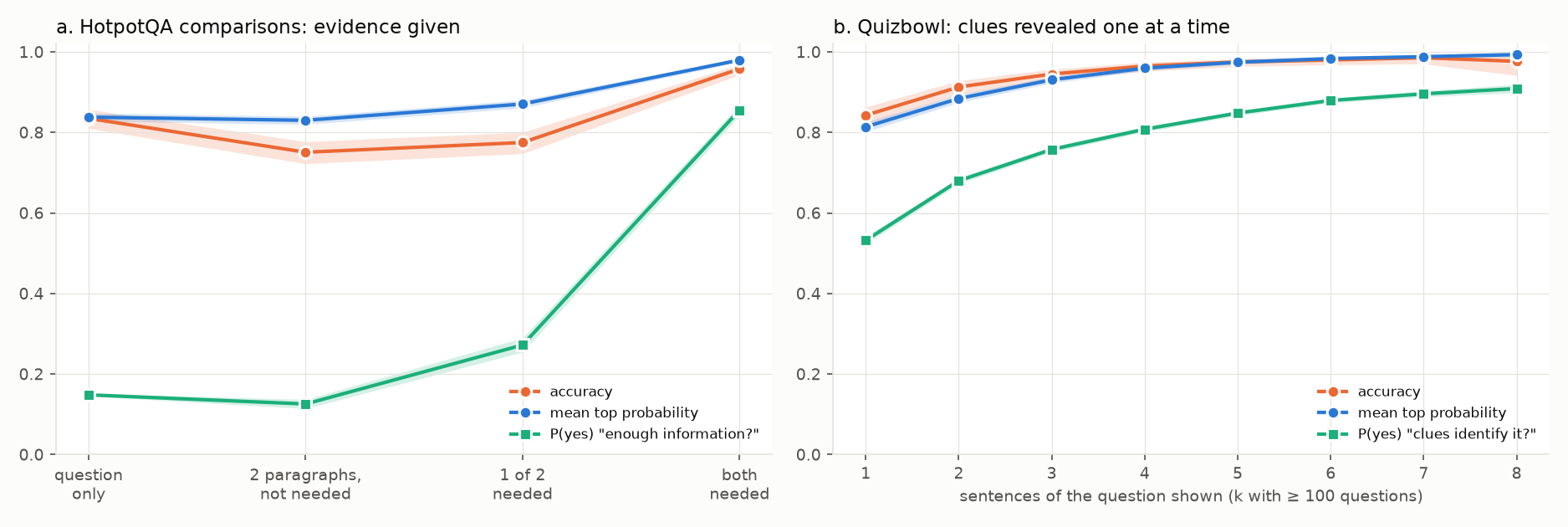}
\caption{Evidence in context. (a) HotpotQA: supporting paragraphs raise
accuracy and confidence together; non-supporting paragraphs reduce accuracy
without a corresponding confidence drop, and partial evidence remains
overconfident; $P(\text{enough})$ rises only with complete evidence. (b) Quizbowl: confidence and accuracy rise together clue by clue.}
\end{figure}

\begin{figure}[h]
\centering
\includegraphics[width=\linewidth]{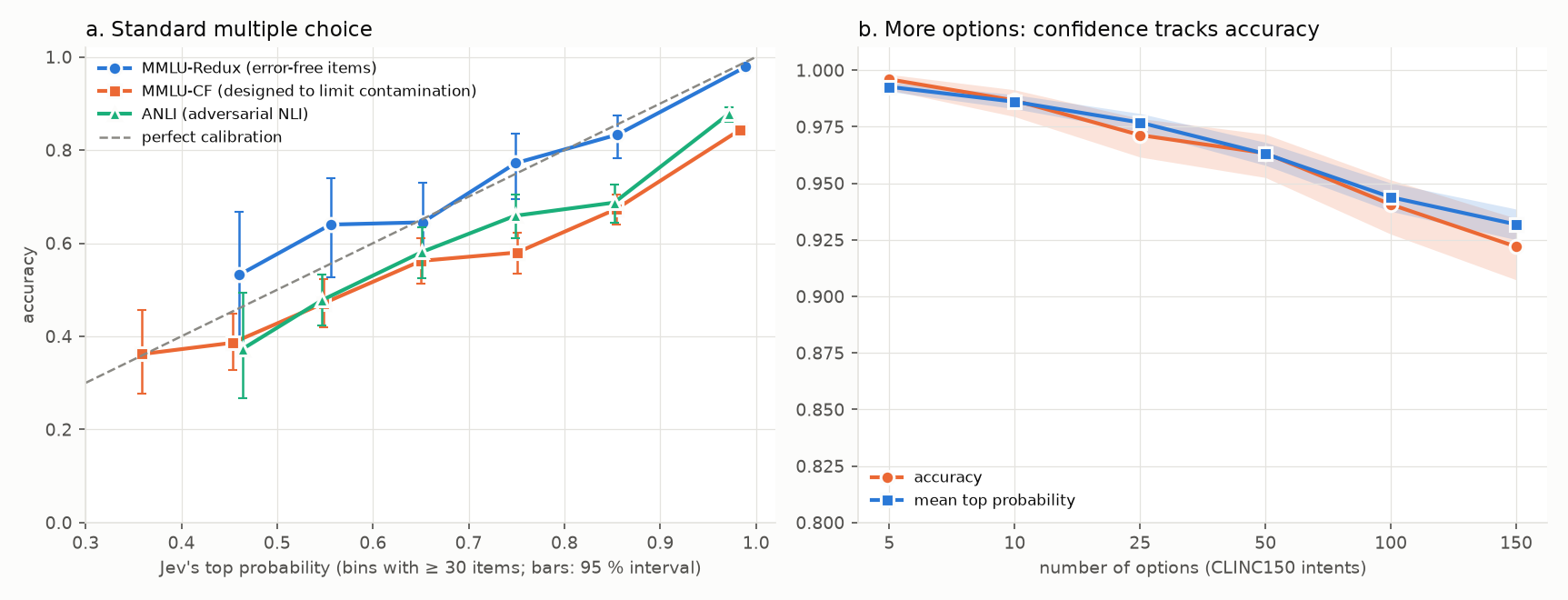}
\caption{Standard multiple choice. (a) Reliability: MMLU-Redux on the
diagonal; MMLU-CF and ANLI below it. (b) CLINC150 with 5--150 options.}
\end{figure}

\begin{table}[h]
\centering\small
\caption{Edge-of-knowledge settings. Gap = mean confidence $-$ accuracy.}
\begin{tabular}{@{}l r r r r r@{}}
\toprule
Setting & $n$ & Accuracy & Confidence & Gap & SmoothECE \\
\midrule
Fabricated entities (uniform reference 0.25) & 1{,}600 & --- & 0.518 & --- & --- \\
Daily Oracle yes/no, after the change point & 1{,}680 & 0.511 & 0.816 & $+0.305$ & 0.305 \\
MMLU-CF & 10{,}000 & 0.776 & 0.914 & $+0.138$ & 0.145 \\
ANLI rounds 1--3 & 3{,}200 & 0.739 & 0.839 & $+0.100$ & 0.105 \\
HotpotQA, non-supporting paragraphs & 951 & 0.751 & 0.830 & $+0.079$ & 0.083 \\
HotpotQA, one of two needed paragraphs & 951 & 0.775 & 0.871 & $+0.096$ & 0.104 \\
Banking77, one example per intent & 770 & 0.681 & 0.828 & $+0.147$ & 0.170 \\
\bottomrule
\end{tabular}
\end{table}

\paragraph{Factual QA and abstention.} Converted to four-option Choice:
TriviaQA accuracy 0.955 (SmoothECE 0.031), SimpleQA Verified 0.753 (0.051),
TruthfulQA binary 0.909 (0.041). Answering only at $\pmax\ge0.9$ keeps 365 of
1{,}000 SimpleQA Verified questions with 1 error. As a verifier of a proposed
TriviaQA answer, Jev's yes/no probability reaches AUROC 0.984 and SmoothECE
0.028.

\paragraph{Unanswerable, ambiguous and disputed items.} $P(\text{answerable})$
separates unanswerable SelfAware questions with AUROC 0.882; $P(\text{single
answer})$ flags ambiguous AmbigQA questions only weakly (0.661). Jev's entropy
is rank-correlated with the entropy of 100 human labels on ChaosNLI
($\rho=0.35$) but strongly compressed.

\paragraph{Out-of-scope intents.} An explicit out-of-scope option flags 89\%
of off-topic CLINC150 requests at 3.5\% false alarms (AUROC 0.982), but only
36\% of messages from held-out Banking77 intents with close relatives in the
schema.

\paragraph{Random devices.} Across all 720 option orders, Jev puts 0.80 on
``one'' for a fair die with number words, 0.74 on ``1'' with digits and 0.62
on ``red'' for colour faces; 1.00 on the 50\% face of a loaded die; 0.97 on 7
for the sum of two dice. Mean probability by display position is flat, and
averaging over orders leaves the distance to uniform unchanged (TV 0.632 vs.\
0.634). With tied stated chances, the choice follows the order in which chances
are mentioned (99.6\%), not the display order.

\paragraph{Inferred settledness.} For 360 made-up events in six domains, each
in a past and a future version differing only in the cue, $P(\text{settled})$
separates past from future with AUROC 1.000 from tense alone (0.73 vs.\ 0.14),
0.996 from dates with today's date given (0.68 vs.\ 0.30), and 0.898 from dates
alone (0.64 vs.\ 0.53): Jev does not place itself in time unless told.

\paragraph{The ``not known'' option across mixes.} With a share $s$ of
post-change questions (period-stratified month-block bootstrap, coverage
matched exactly), the option minus a confidence threshold is $-0.034$
[$-0.045$, $-0.005$] at $s=0.05$, $+0.010$ [$-0.002$, 0.043] at the date panel's
own share $s=0.27$, and $+0.051$ [0.016, 0.092] at $s=0.5$; against a
\texttt{known} threshold it is $+0.018$ to $+0.048$ at every share. Confident
errors nearly vanish with the option (6 of 1{,}191 answers vs.\ 330), but this
count is format-dependent (three-way vs.\ two-way distributions).

\paragraph{Wording robustness.} Fourteen paraphrased, reversed and adversarial
wordings of the three follow-up questions all rank the relevant contrast with
AUROC 0.87--1.00, including adversarial \texttt{settled} cases that pit meaning
against chance or future-tense words; absolute levels vary (for \texttt{known},
real subjects get 0.52--0.79 across wordings). For \texttt{known} these are the
original, cue-bearing contrasts (\cref{app:cues}).

\section{Experiment ledger}
\label{app:ledger}

Generated from the response caches: every paid call is one cached record,
attributed to the first experiment whose rebuilt requests include it. Units
are item $\times$ condition combinations. The cache key includes the replicate
index, so the replicates of a request are always separate paid calls; only
requests identical in content \emph{and} replicate index share a call, which
happens when a later experiment reuses an earlier condition (counted under the
earlier experiment) or when generated items coincide in text. The intent
study's 27{,}240 units are the main arm (2{,}270 items $\times$ 9 conditions,
reported in \cref{tab:intent}) plus two robustness arms on 681 of those items,
alternative example draws (7 conditions) and letter codes instead of numbered
codes (3 conditions). Of the 575{,}442 paid calls, 573{,}234 feed reported
analyses; 2{,}208 are superseded or stray calls that are counted in the cost but
not analysed. Dates are 2026 (UTC).

{\small
\begin{longtable}{@{}p{0.34\textwidth} r r r r p{0.18\textwidth}@{}}
\caption{Items, replicates and paid calls per experiment. $^*$Planned before the
run, plan committed afterwards (not externally verifiable).}\\
\toprule
Experiment & Units & Replicates & Paid calls & Dates & Planning \\
\midrule
\endhead
Instrument checks & --- & varied & 402 & 26 Sep & planned \\
Intent study, development pilot & 6{,}356 & 3 & 7{,}945 & 26 Sep & pilot before the freeze \\
Intent dose--response (test) & 27{,}240 & 3 & 76{,}953 & 26--28 Sep & plan frozen in a git tag \\
Out-of-scope detection & 3{,}570 & 3 formats & 10{,}710 & 28 Sep & planned \\
Knowledge$\times$chance scenarios & 6{,}000 & 3 & 18{,}000 & 28 Sep & planned$^*$ \\
Knowledge boundary (PopQA, fabricated, news) & 26{,}927 & 3 & 80{,}781 & 28 Sep & planned$^*$ \\
Evidence in context (HotpotQA, Quizbowl) & 10{,}622 & 3 & 31{,}866 & 28 Sep & planned$^*$ \\
Standard benchmarks & 22{,}583 & 3 & 67{,}749 & 28 Sep & planned$^*$ \\
Wording robustness & 3{,}080 & 3 & 9{,}231 & 28 Sep & planned$^*$ \\
Odds formats & 1{,}440 & 3 & 4{,}320 & 28 Sep & planned$^*$ \\
Date / ``not known'' option & 13{,}440 & 3 & 30{,}240 & 28 Sep & planned$^*$ \\
``Paragraphs may be irrelevant'' & 2{,}853 & 3 & 8{,}559 & 28 Sep & planned$^*$ \\
Standard multiple choice, option counts & 27{,}900 & 3 & 83{,}604 & 28 Sep & planned$^*$ \\
Second look & 15{,}523 & 3 & 46{,}566 & 28 Sep & planned$^*$ \\
Random devices & 3{,}368 & 1--5 & 4{,}040 & 28 Sep & planned$^*$ \\
Policy cases & 3{,}000 & 3 & 8{,}562 & 28 Sep & planned$^*$ \\
Open benchmark adaptations & 6{,}663 & 3 & 19{,}989 & 28 Sep & planned$^*$ \\
Inferred settledness & 2{,}880 & 3 & 8{,}640 & 29 Sep & written before the run \\
Dates removed & 3{,}360 & 3 & 10{,}077 & 30 Sep & exploratory \\
Look-alike names & 600 & 3 & 1{,}800 & 30 Sep & exploratory \\
Evidence for made-up subjects & 800 & 3 & 2{,}400 & 30 Sep & exploratory \\
Opposite answer proposed & 3{,}360 & 3 & 10{,}080 & 30 Sep & exploratory \\
Random devices, four wordings & 480 & 1 & 480 & 30 Sep & exploratory \\
Negative verification wording & 6{,}720 & 3 & 20{,}160 & 30 Sep & exploratory \\
Today's date just after the event & 3{,}360 & 3 & 10{,}080 & 30 Sep & exploratory \\
Superseded or stray calls & --- & --- & 2{,}208 & 28--30 Sep & not used \\
\midrule
Total & & & 575{,}442 & & \\
\bottomrule
\end{longtable}
}

\section{Datasets}
\label{app:datasets}

\begin{table}[h]
\centering\small
\caption{Public datasets, pinned revisions and licences.}
\begin{tabular}{@{}l l l@{}}
\toprule
Dataset & Source @ revision & Licence \\
\midrule
Banking77 \citep{casanueva2020efficient} & PolyAI task-specific-datasets @ 57ec275d & CC BY 4.0 \\
CLINC150 \citep{larson2019evaluation} & \texttt{clinc/clinc\_oos} @ 155b9c71 & CC BY 3.0 \\
PopQA \citep{mallen2023popqa} & \texttt{akariasai/PopQA} @ 098765c7 & none stated \\
Daily Oracle \citep{dai2025dailyoracle} & \texttt{agentic-learning-ai-lab/daily-oracle} @ 455b35b2 & CC BY 4.0 \\
HotpotQA \citep{yang2018hotpotqa} & \texttt{hotpotqa/hotpot\_qa} @ 1908d6af & CC BY-SA 4.0 \\
Quizbowl \citep{rodriguez2019quizbowl} & \texttt{community-datasets/qanta} @ e3c56022 & unknown \\
SelfAware \citep{yin2023selfaware} & \texttt{ShuoZheLi/SelfAware} @ ffdfba5c & Apache-2.0 \\
SimpleQA Verified \citep{simpleqaverified2025} & \texttt{google/simpleqa-verified} @ 0dc97e0d & MIT \\
TruthfulQA \citep{lin2022truthfulqa} & \texttt{truthfulqa/truthful\_qa} @ 741b8276 & Apache-2.0 \\
TriviaQA \citep{joshi2017triviaqa} & \texttt{mandarjoshi/trivia\_qa} @ 0f7faf33 & unknown \\
AmbigQA \citep{min2020ambigqa} & \texttt{sewon/ambig\_qa} @ e969d013 & CC BY-SA 3.0 \\
ChaosNLI \citep{nie2020chaosnli} & \texttt{earino/chaosnli} @ bf8d66a1 & CC BY-NC 4.0 \\
MMLU-Redux 2.0 \citep{gema2025mmluredux} & \texttt{edinburgh-dawg/mmlu-redux-2.0} @ 372ea425 & CC BY 4.0 \\
MMLU-CF \citep{zhao2024mmlucf} & \texttt{microsoft/MMLU-CF} @ c25b89a9 & CDLA-Permissive-2.0 \\
ANLI \citep{nie2020anli} & \texttt{facebook/anli} @ 8e4813d8 & CC BY-NC 4.0 \\
\bottomrule
\end{tabular}
\end{table}

Free-text benchmarks were converted to four-option Choice: distractors are
other objects of the same relation (PopQA), perturbed dates and numbers or the
most similar answers of the same type (SimpleQA Verified), the most similar
other answers (TriviaQA, Quizbowl), or the dataset's own incorrect answer
(TruthfulQA binary). Options are shown in seeded random order except MMLU,
whose options keep their original order. The ChaosNLI and SelfAware mirrors
were checked against independent copies. The six generated sets are: (1) 480
knowledge$\times$chance scenarios; (2) fabricated-entity questions, 1{,}600
with pseudo-word names and 600 with look-alike names; (3) random devices
(dice, coins, cards); (4) 600 rule-based policy cases; (5) 360 made-up events
for inferred settledness; and (6) adversarial scenarios for the wording
robustness of the follow-up questions.

\end{document}